\documentclass[letterpaper,12pt]{article} 
\usepackage[draft]{hyperref} 
\RequirePackage{geometry}
\usepackage{cite}
\usepackage{color}
\usepackage[pdftex]{graphicx}
\usepackage[small,bf]{caption}
\usepackage{subcaption}
\usepackage{verbatim} 
\usepackage{amssymb,amsmath,amsthm}
\usepackage[parfill]{parskip}
\usepackage{enumerate}

\newcommand{\bb}{\mathbf}
\usepackage{tikz}
\usetikzlibrary{arrows}
\usepackage{pgfplots}
\title{Imaging with Rays: Microscopy, Medical Imaging, and Computer Vision}
\author{Keith Dillon and Yeshaiahu Fainman}

\begin{document}
\maketitle
\begin{abstract}
In this paper we broadly consider techniques which utilize projections on rays for data collection, with particular emphasis on optical techniques.
We formulate a variety of imaging techniques as either special cases or extensions of tomographic reconstruction.
We then consider how the techniques must be extended to describe objects containing occlusion, as with a self-occluding opaque object. 
We formulate the reconstruction problem as a regularized nonlinear optimization problem to simultaneously solve for object brightness and attenuation, where the attenuation can become infinite.
We demonstrate various simulated examples for imaging opaque objects, including sparse point sources, a conventional multiview reconstruction technique, and a super-resolving technique which exploits occlusion to resolve an image.
\end{abstract}



\section{Introduction}
The primary motivating research question here is how do we formulate an imaging system that can reconstruct ``anything'' given a collection of views. 
We consider the high-frequency limit wherein we can formulate a variety of imaging techniques as either special cases or extensions of tomographic reconstruction.
The unifying factor is that they all collect projections along rays. 
An important benefit of this approximation is the resulting sparsity of the matrices describing the system which allow large-scale computations to be performed more efficiently. 
And a tomographic formulation of image formation can cover a very broad range of applications.
But there is are important limitations when the illumination is at optical frequencies, which leads us to the question of how to deal with occlusion in such systems.
This in turn leads to ideas for a variety of related imaging systems, and the motivation for digging deeper into understanding the data-dependent performance of modern imaging techniques.

\section{Rays and Projections}

\begin{figure}[!h]\centering 
    \input{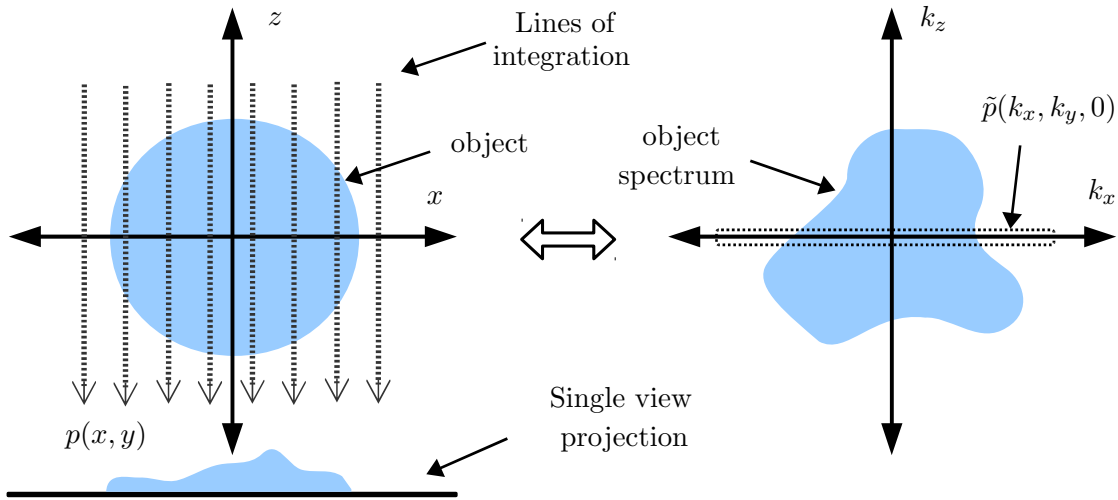} 
    \caption{Tomographic projection and $k$-space. $z$ is the vertical axis and $x-y$ is the perpendicular plane, cutting into the page at the horizontal line depicted.} 
    \label{tomo_projection_1}
\end{figure}

In this section we give examples of the breadth of applications of techniques which reconstruct the object from projections along rays, including most of the major imaging modalities, and we discuss the structure such systems have in common.
A projection of an object along the $z$-axis as depicted in Fig. \ref{tomo_projection_1} is written as  
\begin{align}
  p(x,y) = \int f(x,z,y) dz
\end{align}
Integrals here run over the entire real axis, generally with an object of limited size (i.e $f(x,y,z)$ is zero outside some finite region).
By taking the Fourier transform of the projection we get
\begin{align}
  \tilde{p}(k_x,k_y) &= \int \int \left[ \int f(x,z,y) dz \right] \text{e}^{k_x x+k_y y} dx dy \notag \\
                       &= \int \int \int f(x,z,y) \text{e}^{k_x x+k_y y + 0 z} dx dy dz \notag \\
                       &= \tilde{f}(k_x, k_y, 0)
\end{align}
The result, $\tilde{f}(k_x, k_y, 0)$, is a slice of the Fourier transform of $f(x,y,z)$. 
By rotating the direction of the projection we can similarly get slices along rotated planes (most easily seen by simply rotating the coordinate system).
This is the projection-slice theorem \cite{deans_radon_2007}.

The Fourier domain, referred to as $k$-space in the medical imaging field, is useful to visualize the data collection and requirements for reconstruction.
Using signal processing theory, we would expect to need a collection of samples covering a $k$-space region (in every dimension) up to some maximum spatial frequency determined by the spectrum of the object, and with a sample density in $k$-space sufficient to cover the extent of the spatial size of the object in $(x,y,z)$.
By resampling the $k$-space samples to an appropriate grid the object may be reconstructed by so-called direct Fourier techniques utilizing fast Fourier transforms. 

A complete collection of projections at all angles (up some some desired sample density and size) is not always possible, though it is mathematically necessary to guarantee a unique solution. 

\begin{figure}[!h]\centering 
    \scalebox{0.42}{\includegraphics[trim=0in 7.1in 0in 0.0in]{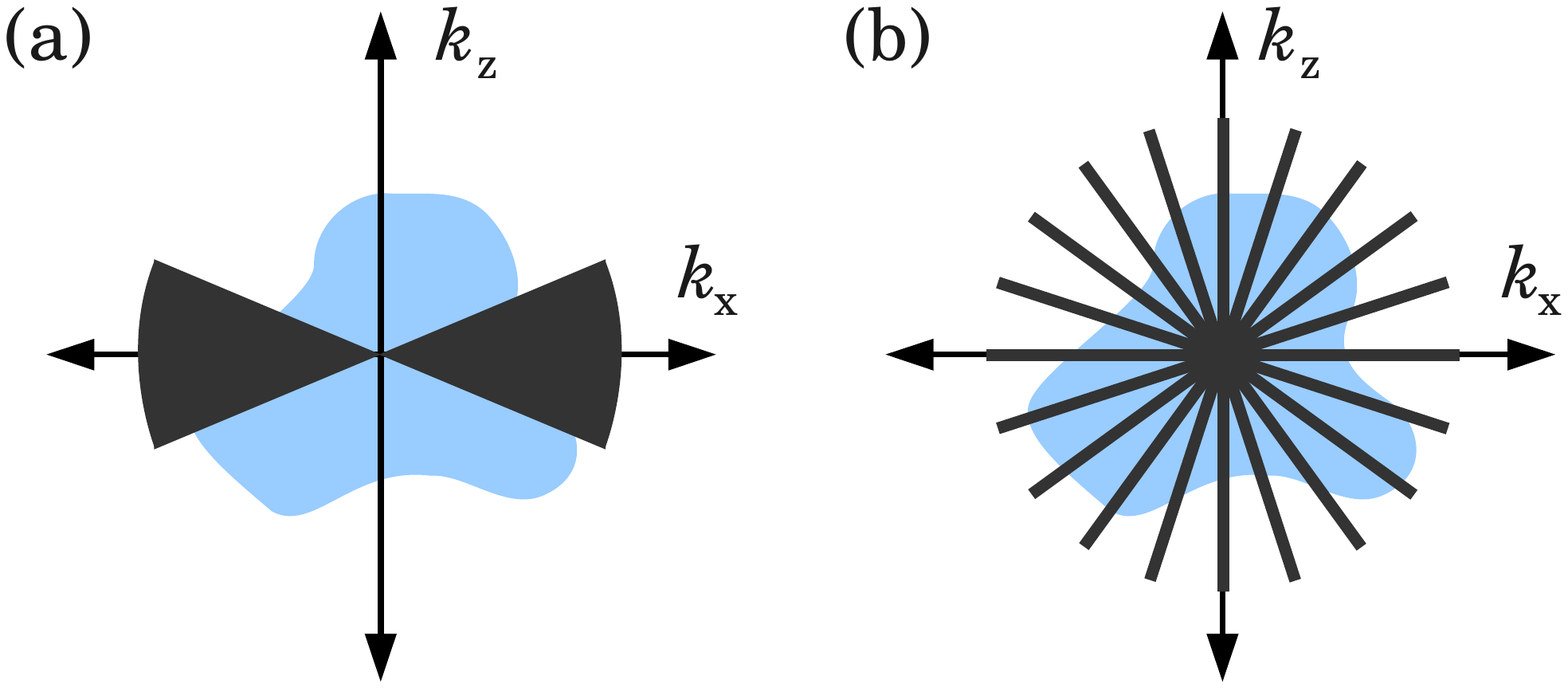}} 
    \scalebox{0.42}{\includegraphics[trim=0in 7.1in 0in 0.0in]{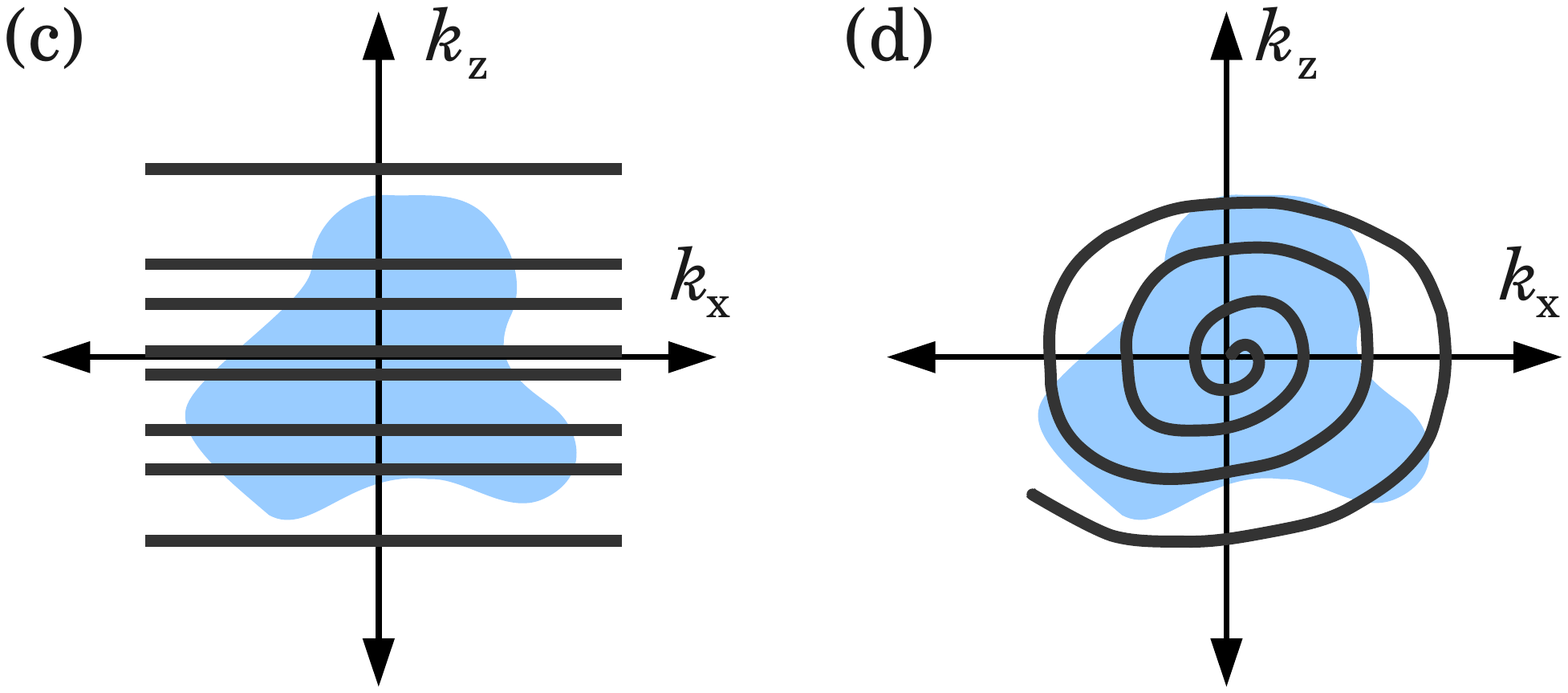}} 
    \caption{Limited angle collection, sparsely spaced projections, then arbitrary $k$-space sampling examples.} 
    \label{tomo_projection_2}
\end{figure}

Examples of the $k$-space regions collected for some incomplete collections is shown in Fig. \ref{tomo_projection_2}, where the gray areas represent regions where samples are collected.
Fig. \ref{tomo_projection_2}(a) depicts a limited-angle case, a well-known problem in tomography \cite{davison_ill-conditioned_1983}. 
Fig. \ref{tomo_projection_2}(b) depicts a sparse collection, often used as an example for how the prior information mentioned above can allow for reconstruction with incomplete data. 
These would both be collected with systems utilizing parallel-beam projections.
Figs. \ref{tomo_projection_2}(c) and (d) depict examples of the arbitrary $k$-space sampling possible with magnetic-resonance imaging (MRI) systems, which in essence, perform direct sampling of $k$-space rather than indirectly via projections along rays.

\begin{figure}[!h]\centering 
    \includegraphics[trim=0in 4.2in 0in 0in]{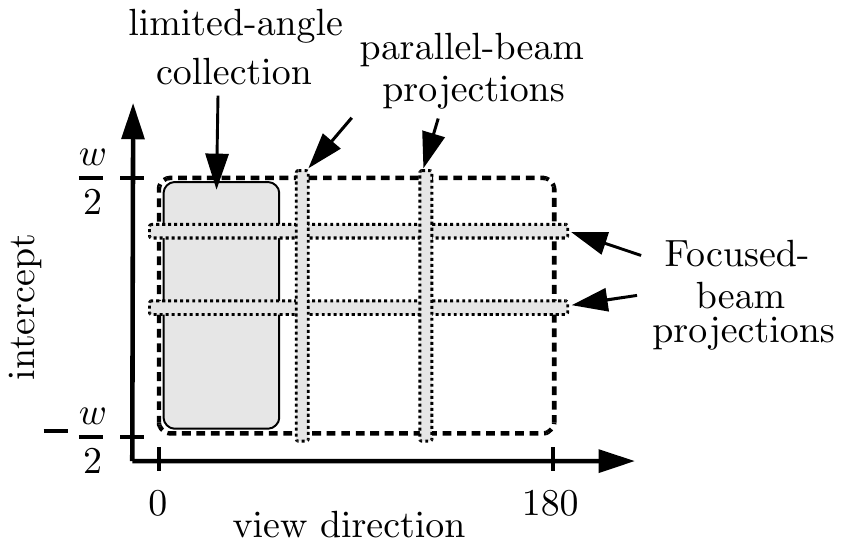} 
    \caption{Slant-stack region necessary for reconstruction, and regions collected by different projections.} 
    \label{radon_1}
\end{figure}

Another way to visualize the data collection is depicted in Fig. \ref{radon_1}, sometimes known as the slant stack.
In the medical imaging field it is also referred to as a sinogram, since a point on the object will follow a sinusoidal path through the slant stack for a conventional computed tomography (CT) collection, though some use different terms to differentiate between continuous and discrete versions of the information \cite{averbuch_fast_2001}.
It may also have the axes transposed from the representation used here (with view angle on the vertical axis).
In the slant stack the value of each point represents a projection integrated along a single ray, with the coordinates giving the position and direction of the ray.
This is a version of the Radon transform of the object \cite{deans_radon_2007}, which can be written as Eq. (\ref{radon_eqn_0}).
\begin{align}
  g(r,\theta) = \int f(x, r+ x \cos \theta) dx 
\label{radon_eqn_0}
\end{align}
Here, $f(x,z)$ is a slice of the object, i.e, for a fixed choice of $y$.

This perspective can be useful (particularly in two dimensions) to understand how more arbitrary collections of projections can be related to the $k$-space picture. 
A parallel-beam projection produces a vertical slice of the slant stack as depicted in Fig. \ref{radon_1}.
Other types of projections (i.e. for which each view is some set of projections along non-parallel rays) produce some other set of points within the slant stack. 
And while non-parallel collections may not be easy to relate to $k$-space individually, a sufficient collection of them may produce filled vertical lines which may then be Fourier-transformed to utilize the projection-slice theorem as before.
Notably, a diverging or converging (i.e. focused) beam may, within the proper choice of coordinates, produce horizontal slices of the slant stack as depicted in Fig. \ref{radon_1}.





\subsection{Computed Tomography}

Computed tomography (CT) is the most basic projection-imaging technique.
CT systems typically utilize an x-ray source which is used to project radiation through the object for multiple views for a full sweep of angles around a patient \cite{buzug_computed_2008}.
The parallel-beam formulation described earlier was initially used as an approximation to the geometry of the rays emanating from the source. 
More accurate formulations are based on a so-called cone beam, describing diverging rays. 
Analogous systems have also been demonstrated with light at optical and infrared frequencies \cite{zysk_projected_2003, lue_synthetic_2008, marks_cone-beam_2001,marks_three-dimensional_1998,dillon_computational_2010}.
The optical case relies on the Eikonal approximation \cite{born_principles_2000}, which models the high frequency limit where light rays behave as x-rays.

A general expression for the projection along a ray can be written as Eq. (\ref{generalray}), where we have included the refractive index variations for the optical cases \cite{dillon_computational_2009}.
\begin{align}
   p(x,y) &= \exp\left\{-\int_{\bb r_0}^{\bb r_1} \alpha(\bb r) ds + j k \int_{\bb r_0}^{\bb r_1} \Delta n(\bb r) ds \right\} 
\label{generalray}
\end{align}
The integrals follow the ray path from $\bb r_0$ to $\bb r_1$, points on the source and detector, respectively, and $\bb r$ implies some point $(x,y,z)$.
$\alpha(\bb r)$ is the spatially-varying attenuation coefficient and $\Delta n(\bb r)$ is a spatially-varying refractive index.
These are the parameters to be reconstructed.
We combine these into $\bb \eta(\bb r) = \alpha(\bb r) -jk \Delta n(\bb r)$.
We can write this projection as Eq. (\ref{generalray2}), where the integrals run over all space.
\begin{align}
   p(x,y) = \exp\left\{-\int\int\int \eta(\bb r) \delta^2\left(\bb r \in L(\bb r_0, \bb r_1)\right) d\bb r  \right\} 
\label{generalray2}
\end{align}
$L(\mathbf r_0, \mathbf r_1)$ is the set of points on the line segment between $\mathbf r_0$  and $\mathbf r_1$.
In the case where the ray is vertical for example, it can be parametrized by points on the vector $(x_1,y_1,z)^T$, so $\delta^2\left(\bb r \in L(\bb r_0, \bb r_1)\right) = \delta(x-x_1)\delta(y-y_1)$.
And using the sifting property of the delta function we get Eq. (\ref{generalray3}).
\begin{align}
   p(x_1, y_1) = \exp\left\{-\int \eta(x_1, y_1, z)  d z  \right\} 
\label{generalray3}
\end{align}
If we instead parametrize the ray at some angle, such as with the vector $(x_1+z \cos\theta_x,y_1+z \cos\theta_y ,z)^T$, we get Eq. (\ref{generalray4}), where we have also taken the log of both sides.
\begin{align}
   \log p_{\theta_x,\theta_y}(x_1, y_1) = -\int \eta(x_1+z \cos\theta_x,y_1+ z \cos\theta_y,z) d z  
\label{generalray4}
\end{align}
Here we have a version of the Radon transform, but the system is not linear in the unknown parameters (i.e. the relationship between the unknown and the data is not a linear system), rather it is logarithmic.
Versions of tomography that result in systems like this are referred to as transmission tomography.

Eq. (\ref{generalray2})  may be written as, after taking the log and generalizing notation somewhat, 
\begin{align}
   \log p(\bb r ) &= -\int\int\int \eta(\bb r') \delta^2\left(\bb r' \in L(\bb r_0, \bb r_1)\right) d\bb r' \\
               &=  \int\int\int  P(\bb r, \bb r') \eta(\bb r')d\bb r'. 
\label{generalray2_1}
\end{align}
Where we have defined a linear operator $P(\bb r, \bb r')$ to describe the system.

\subsection{Magnetic Resonance Imaging}

Magnetic Resonance Imaging (MRI) is an extremely powerful imaging modality which can collect a large variety of information from an object\cite{reiser_magnetic_2007}.
Its operation is quite different than the other modalities in this section but it does allow lend itself to an analogous $k$-space perspective, so we will very briefly summarize it here.
Image formation in MRI is based on a physical effect whereby nuclei in a strong magnetic field become resonant at a characteristic frequency that depends (linearly) on the magnetic field strength.
In very simplified terms, the classical MRI imaging approach employs three gradient magnetic fields, one each for slice encoding, frequency encoding, and phase encoding. 
These gradient fields are designed to be linearly varying  along a single chosen axis, and constant on the remaining axes.

First, a slice encoding gradient field is typically applied while a narrowband electromagnetic signal is applied to the object for a limited time.
Due to the frequency-dependence of the resonance, only a plane (or slice) of the object is excited by this signal, where the field strength is such that the resonant frequency equals the narrowband excitation signal.
If an antenna were to subsequently read the electromagnetic signals re-emitted by the nuclei, it would in effect record a projection of the nuclei density on a plane, and one could perform a three-dimensional Radon transform of the object by collecting many such slices.

Alternatively, by applying a new gradient field during the collection time such that the magnetic field varies linearly across the slice, the nuclei can be made to re-emit their energy at different frequencies.
Then our collected data would consist of projections along parallel rays, each at a different temporal frequency which may be separated using signal processing techniques. 
By rotating this frequency encoding gradient and collecting a stack of measurements, we can form a Radon transform of the slice.
This is known as a projection reconstruction technique \cite{glover_projection_1992}.

However we may also make use of a so-called phase encoding gradient, turned on for a limited interval prior to data collection, to cause a linear phase change across the slice in a direction perpendicular to that to be used in the readout gradient. 
By varying the slope of this phase gradient over many collections, we can apply a frequency modulation which varies across the slice, allowing us to reconstruct the object in this final coordinate using a Fourier transform in that coordinate as well.
Variations on the above basic process can be employed to provide arbitrary collection patterns for $k$-space samples.

\subsection{PET and SPECT}

In positron-emission tomography (PET) and single-photon emission CT (SPECT), a radioactive isotope known as a tracer is injected into a patient.
The tracer acts as a marker, accumulating at locations of interest. 
As the isotope decays, it will ultimately produce photons that can be collected at a detector, using a technique that is closely related to CT \cite{lewitt_overview_2003}.
Mathematically, a collection of such data (using a technique analogous to CT) can be formulated as a so-called attenuated x-ray transform, which we may write as (see later sections for a derivation of a similar system)
\begin{align}
  g(r,\theta) = \int \exp\left\{-\int \alpha \left(x, r+ x \cos \theta\right) \right\} s(x, r+ x \cos \theta) dx. 
\label{xray0}
\end{align}
Here, $\alpha(x,z)$ is a two-dimensional attenuation map and $s(x,z)$ is the emission or brightness map.
Depending on the choice of parametrization, variations of this are also referred to as the attenuated Radon transform \cite{gullberg_attenuated_1979}.
Inversion techniques have been demonstrated for such problems when the attenuation is known \cite{natterer_inversion_2001, novikov_inversion_2002}.
In practice transmissive measurements (i.e. conventional CT) are collected at the same time as the collection of the emission data \cite{zaidi_determination_2003}.

While the underlying physics of the system remains linear, the inverse problem of estimating the object parameters (i.e. the relationship between the unknowns and the measured data) becomes nonlinear when the attenuation is also unknown \cite{dillon_nonlinear_2011}.
This is also referred to as the identification problem, \cite{finch_attenuated_2003, stefanov_identification_2011}.
We will consider a related system in more detail later when we address opacity.

\subsection{Transmissive Microscopy}

Microscopy systems operate at the small-size extreme of imaging techniques. 
The optical system is much larger than the object, and the numerical aperture and therefore range of collection angles for a single view is typically very large.

\begin{figure}[!htbp]\centering 
    \includegraphics[trim=0in 3.1in 0.0in .0in]{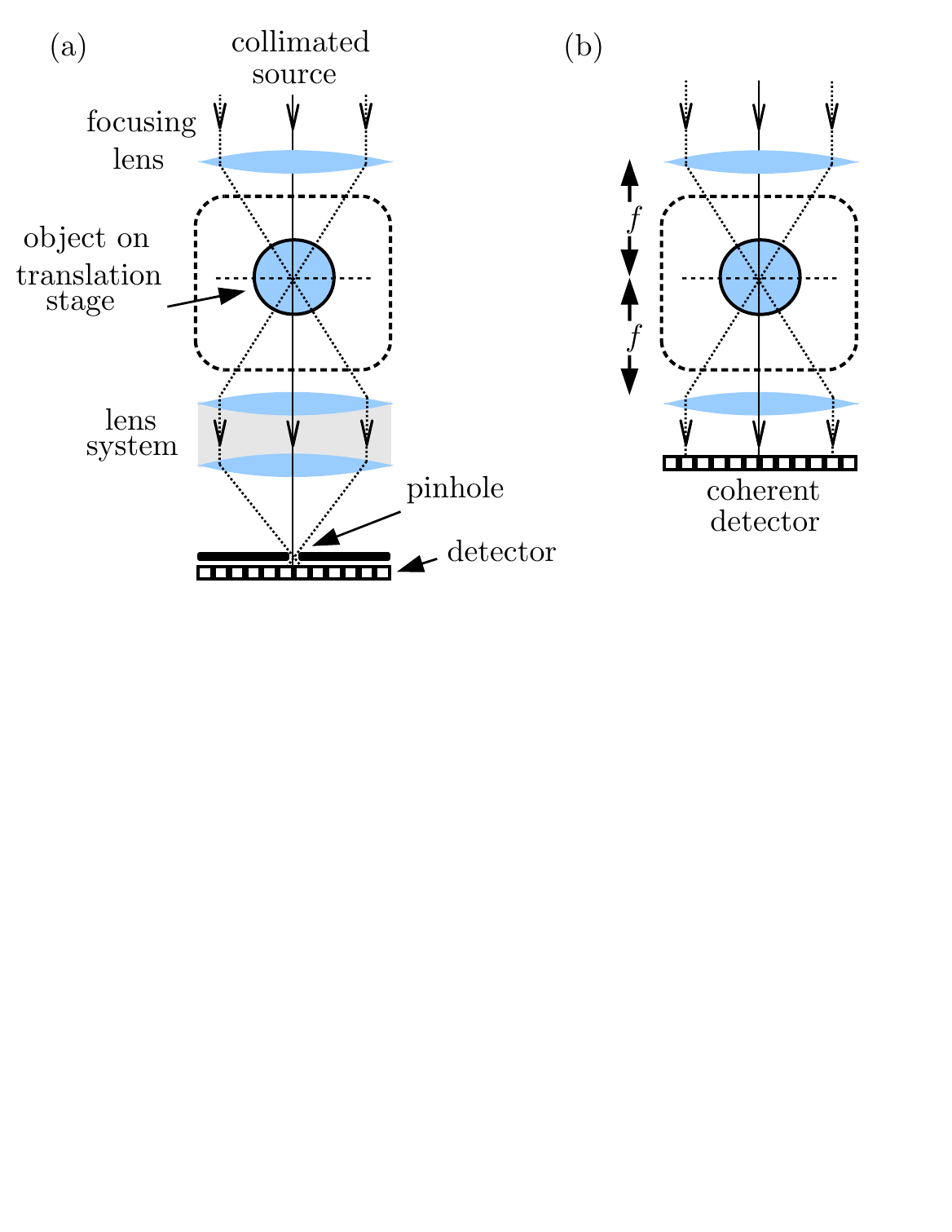} 
    \caption{Conventional and computational transmissive confocal microscope} 
    \label{confocal_tomo}
\end{figure}
The conventional confocal microscope is shown in Fig. \ref{confocal_tomo}(a), in a transmissive mode.
We will consider the variation in Fig. \ref{confocal_tomo}(b), known as a computational confocal \cite{dillon_computational_2009} or digital confocal \cite{goy_multiple_2013} microscope.
If $z$ is the vertical direction, $u(x',y')$ is the field at the plane of the detector, and $(x_0,y_0,z_0)$ are the coordinates of the focus, we have the following expression for $u$
\begin{align}
    u(x',y';x_0,y_0,z_0) 
    &= \exp\left\{\int \eta \left( \frac{x'}{f}(z-z_0)+x_0, \frac{y'}{f}(z-z_0)+y_0 , z \right) dz \right\}.
    \label{confocal_radon_1}
\end{align}    
By taking the log of the measured signal we again have a version of the Radon transform. 
In this case if we take multiple measurements by scanning in the $x$ and $y$ directions (by varying $x_0$ and $y_0$), we would collect only a limited-angle region of the slant stack as determined by the numerical aperture of the optical system.
This variant of the Radon transform can be interpreted as performing a shearing operation rather than a rotation. 
Rays at different angles transverse the same distance in the $z$-direction, but transverse increasing distances the farther they are from the axial.


In this particular system, we can find a simple relation to reconstruct the focal plane of the object.
Commonly, the goal of a confocal microscopy system is to image a selected depth section of the object, perhaps in real time so that biological processes may be observed.
The focal plane of a computational confocal microscope may be computed efficiently using the projection-slice theorem \cite{dillon_depth_2010} (see Appendix C), as 
\begin{align}
    \left\{ \int \int \log u(x';y';x_0,y_0,z_0) dx' dy' \right\} \ast h(x_0, y_0) = 2 \pi \eta (x_0,y_0,z_0)
\end{align}    
Where ``$\ast$'' describes the convolution operation and $h(x_0, y_0)$ is a convolution kernel performing a high-pass filtering with respect to the scan direction. 
This expression says we simply need to integrate over the entire signal amplitude at the detector for each point in the scan, then apply a high pass filtering over successive steps in the scan of these scalar integrated values.

Generally we can write Eq. (\ref{confocal_radon_1}) as 
\begin{align}
    \log u(\bb r) 
    &= \int \int \int P(\bb r, \bb r'; z_0) \eta \left( \bb r' \right) d \bb r', 
    \label{confocal_P0}
\end{align}    
Where
\begin{align}
  P(\bb r, \bb r'; z_0) = \delta^2 \left( \frac{x}{f}(z'-z_0)+x - x', \frac{y}{f}(z'-z_0)+y -y' , z' \right). 
    \label{confocal_P1}
\end{align}

\subsection{3D Photography}
Computational photography techniques operate at the opposite extreme from microscopy, involving large distances and small aperture angles.
Related techniques include multi-view reconstruction in computer vision, where it is also called volumetric reconstruction \cite{dyer_volumetric_2001}. 
The goal there is also to use multiple two-dimension images (e.g. photographs) from different views, to reconstruct a three-dimensional object.  
This might be desired directly for the formation of computer models \cite{pollefeys_images_2002}, or for applications such as compression for 3D video \cite{kubota_multiview_2007}.
The most common approaches used tend to be heuristic techniques such as voxel coloring 
\cite{seitz_photorealistic_1999}, 
which presuppose the object consists of opaque points, and try to find their locations.

The pinhole camera is a geometrical optics approximation to a basic camera imaging system.
\begin{figure}[h!] \centering 
    \includegraphics[trim=0.0in 3.3in 0.0in 0.0in]{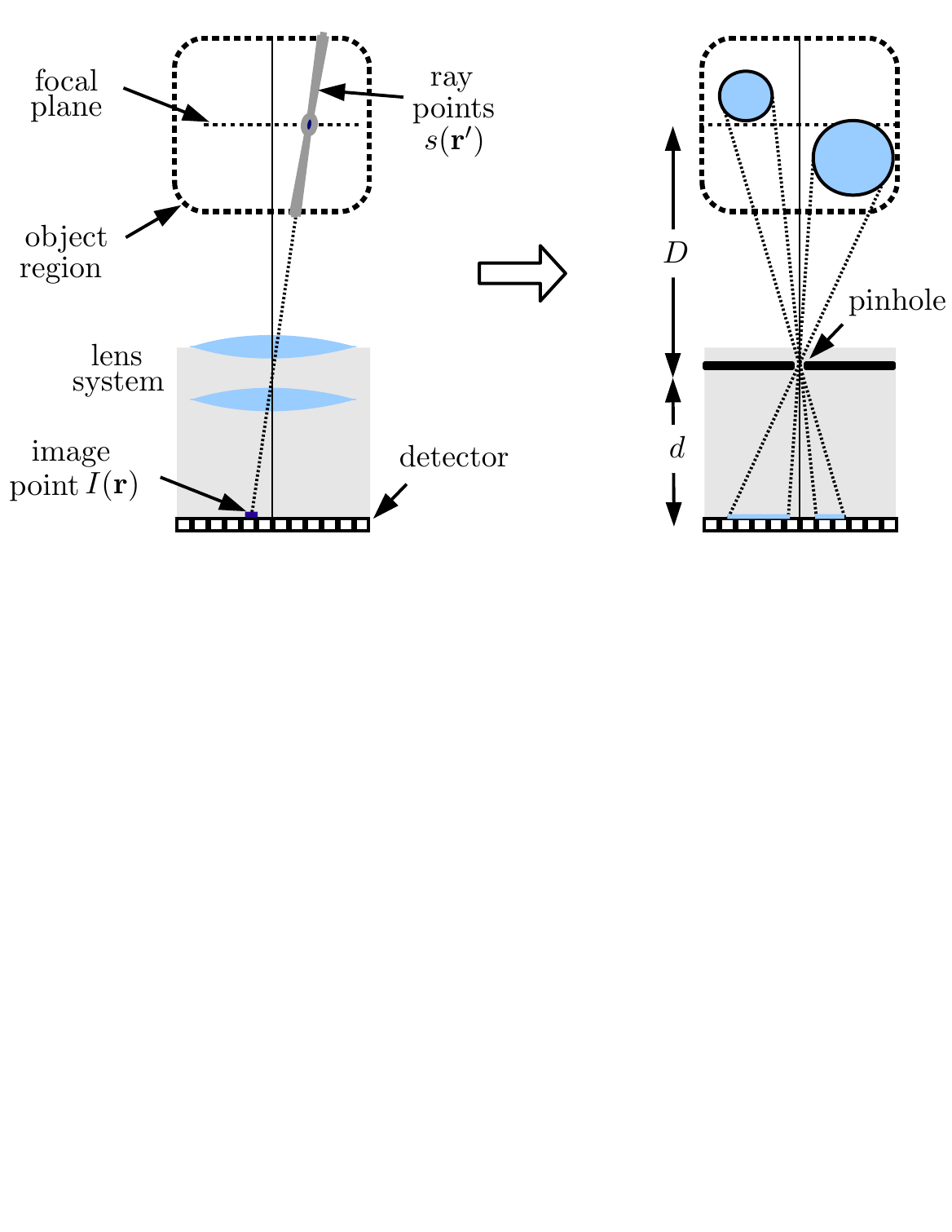} 
    \caption{Simple camera model approximated as a pinhole camera.} 
    \label{opticalattenuatedradon_pinhole3}
\end{figure}
In the camera model of Fig. \ref{opticalattenuatedradon_pinhole3}
we take the $z$-direction to be the vertical axis, so the detector and focal plane are parallel to the $x$-$y$ plane.
If we further assume the pinhole is at the origin, then in the geometrical optics approximation the light collected at a point on the detector will be
\begin{equation} \label{eq:pinhole1}
  \begin{split}
    I (x,y,-d) &= \int_{D-L}^{D+L} s \left( \frac{x}{d}z^\prime, \frac{y}{d} z^\prime, z^\prime \right) dz^\prime \\
  \end{split}
\end{equation}
Where we have defined the size of the object region to be $L$. 
The constant of proportionality is based on the camera and detector characteristics.
Note that we needed to neglect occlusion for the detector point to collect all the light along the ray.
We can write Eq. (\ref{eq:pinhole1}) as
\begin{equation} \label{eq:pinhole2}
  \begin{split}
    I (x,y,-d) &= \int\int\int \delta^2 \left( x^\prime - \frac{x}{d}z^\prime, y^\prime - \frac{y}{d} z^\prime, z^\prime \right) s(x^\prime, y^\prime, z^\prime) dx^\prime dy^\prime dz^\prime \\
   &= \int\int\int P (\mathbf r, \mathbf r^\prime) s (\mathbf r^\prime) d \mathbf r^\prime
  \end{split}
\end{equation}
Defining another variation of the operator $P (\mathbf r, \mathbf r^\prime)$.

The above example of $P (\mathbf r, \mathbf r^\prime)$ is based on a single image collected from a single viewpoint. 
In general one might collect images from a variety of views, at once or sequentially, and form a large operator by concatenating the individual  $P (\mathbf r, \mathbf r^\prime)$ operators for all the views together.
Also note that this form of projection is mathematically similar to cone-beam tomography.
If the object is relatively small and far from the camera, the rays will be approximately parallel and multiple views will be mathematically equivalent to multiple tomographic projections with parallel rays (neglecting occlusion).
So we would have, for the projection operator,
\begin{equation} 
  \begin{split}
    P (\mathbf r, \mathbf r^\prime) &= \delta^2 \left( x^\prime - \frac{x}{d}z^\prime, y^\prime - \frac{y}{d} z^\prime, z^\prime \right) \\
   & \approx \delta^2 \left( x^\prime - \frac{x}{d}D, y^\prime - \frac{y}{d} D, z^\prime \right) 
  \end{split}
  \label{eq:pinhole3}
\end{equation}
And the integral of Eq. (\ref{eq:pinhole2}) would collect a projection of parallel rays in the axial direction.

Note that this variation of tomography is linear.
Tomography systems that yield linear systems such as this may be referred to as emission tomography, since the system passively collects radiation of some kind.

Also note that it is not possible to make a linear system like Eq. (\ref{eq:pinhole2}) which describes an object with opacity, as the opacity would violate superposition; 
the output signal from an occluded region is not added to the output signal from the occluding region.



\subsection{$\bb A \bb x = \bb b$, Tomography-Style}

\begin{figure}[h]\centering 
    \input{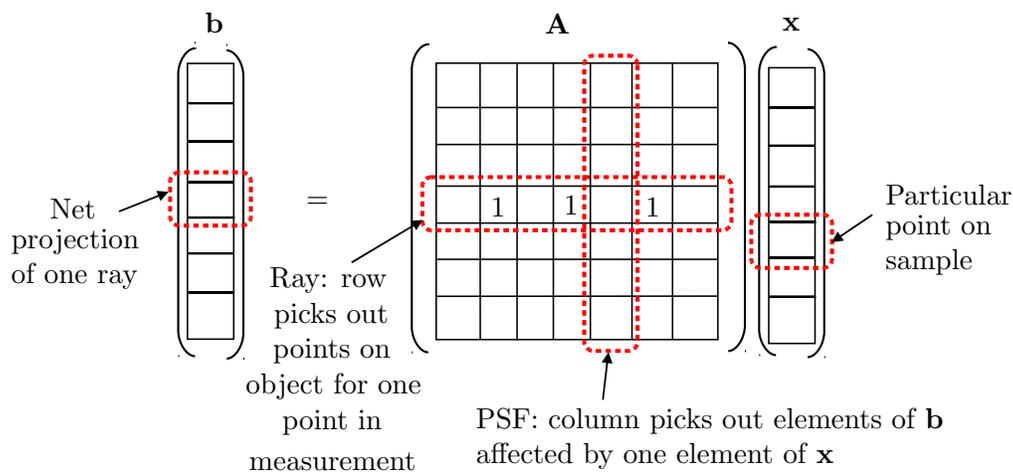} 
    \caption{Input $\bb x$ to forward model as a vector of pixels on the object, and output $\bb b$ of model as a vector of detector samples. If the relationship between $\bb x$ and $\bb b$ is linear, it is described by the matrix $\bb A$. In the inverse problem we are concerned with estimating $\bb x$ given $\bb b$ and $\bb A$.} 
    \label{Axbfig}
\end{figure}

For the linear systems described above, we have an operator kernel $P(\bb r, \bb r')$ that may be discretized to make a linear system of equations. 
Then rather than use techniques such as the inverse Radon transform that rely on the system structure, we can formulate reconstruction as solution of this linear system of equations.
We may view each ray projection as a measurement on a basis function, which we approximate with a sampled version of the function at sufficiently-high resolution.
As a result we have a set of measurements using what might be broadly described as basis vectors, which make the rows of the matrix $\bb A$ in Fig. \ref{Axbfig}. 

The matrix $\bb A$ describes the operator which applies the kernel $P(\bb r,\bb r')$.
The matrix $\bb b$ may denote either the collected data itself, as in the case of emission tomography, or its log, as in the case of transmissive tomography.
And the vector $\bb x$ is a sampled version of the unknown, which may be the attenuation or brightness maps.

So for the linear systems described above, we can easily form the linear system of equations $\bb A \bb x = \bb b$ describing the data collection. 
To estimate the object we must find the vector $\bb x$ consistent with this matrix system.
If we have a collection that does not provide sufficiently-dense sampling over a sufficiently large region of $k$-space to cover the entire spectrum of the object, then the matrix system we formulate to describe this data collection will be underdetermined.


\section{Unknown Opacity}

The use of optics for projection imaging quickly leads to the real-world problem of opacity and occlusion.
A typical 3D photography scene might consist of several opaque objects with empty space between and surrounding them.
As noted earlier, occlusion cannot be described with a linear object-data relationship (i.e. a linear forward model). 
In fact it implies a combination of the transmissive (logarithmic) and emission (linear) effects discussed earlier, and might be viewed as a variation on PET reconstruction with an unknown attenuation map.
The key point however is that the attenuation can become infinite.

\subsection{Combined Source and Attenuation Tomography}

\begin{figure}[h!] \centering 
    \includegraphics[trim=1.0in 7.0in 0.0in 0.0in]{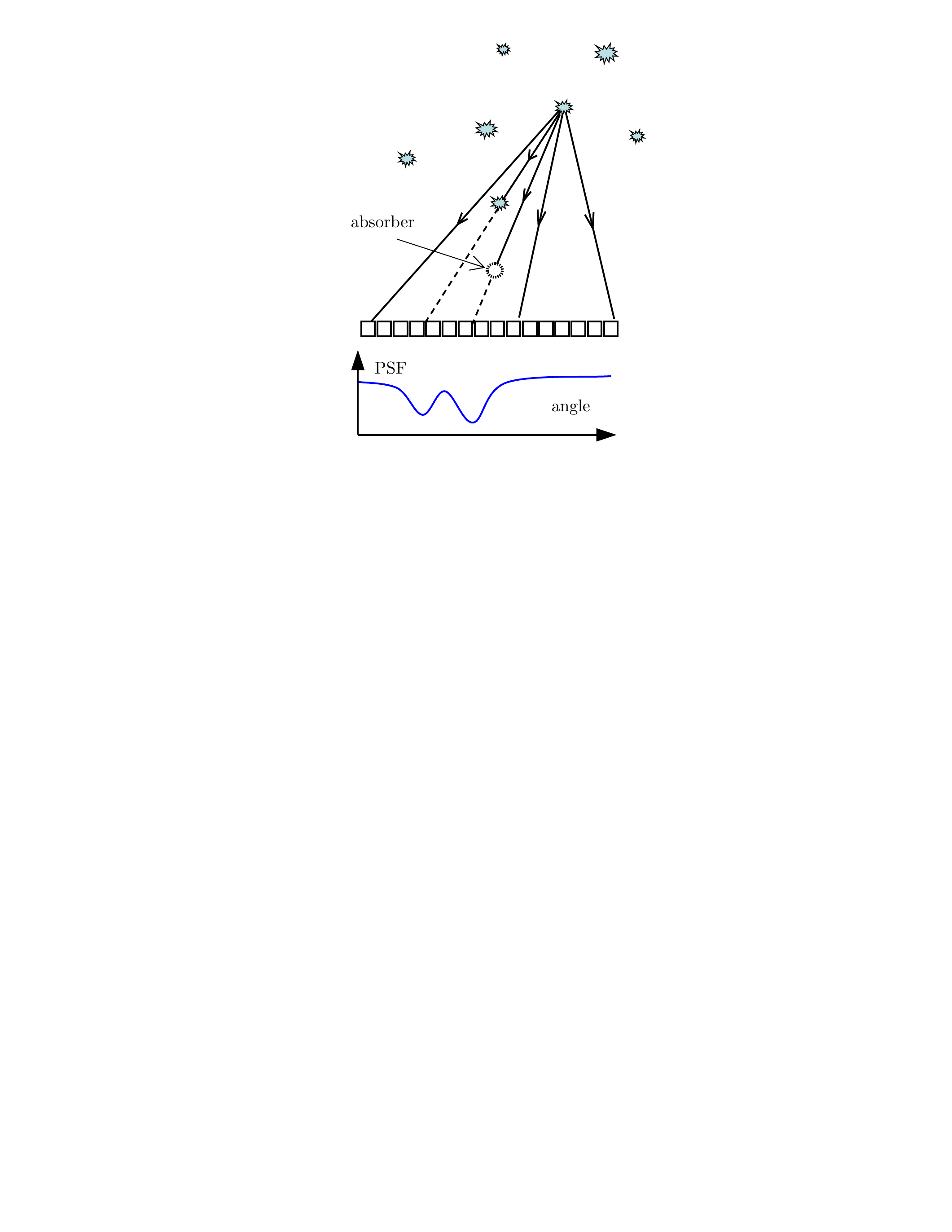} 
    \caption{Object-dependent point-spread function. If we consider the light scattering from a given point over different directions, we get a point-spread function that contains information about subsequent scatterers and absorbers in the region between the point and the detector.} 
    \label{fig:objectdependentpsf}
\end{figure}

If we assume a passive imaging system such as in Fig. \ref{opticalattenuatedradon_pinhole3}, where the emission sources viewed may be the true sources, or may simply be the last scatterer the photon encounter prior to detection,
then we could treat the measured signal as the output of the linear system of Eq. (\ref{eq:green-incoherent2}).
\begin{equation} \label{eq:green-incoherent2}
  \begin{split}
    I ( \mathbf r ) = \int_\Omega H (\mathbf r, \mathbf r^\prime) s (\mathbf r^\prime) d \mathbf r^\prime
  \end{split}
\end{equation}
The operator kernel  $H (\mathbf r, \mathbf r^\prime)$ may be thought of as a spatially-variant point-spread function (PSF), as depicted in Fig. \ref{fig:objectdependentpsf}. 
In our case it is not completely known, as it depends on the object itself.
$s (\mathbf r^\prime)$ is the unknown source brightness, which in the case of multiple-scattering refers to the brightness of the last point from which a photon scattered before being detected. 
$I (\mathbf r)$  is the measured signal intensity.
$\Omega$ is the region containing the unknown object. 

We can describe $H (\mathbf r, \mathbf r^\prime)$  as the product of two terms
\begin{equation} \label{eq:psf1}
  \begin{split}
    H (\mathbf r, \mathbf r^\prime) = P (\mathbf r, \mathbf r^\prime) Q (\mathbf r, \mathbf r^\prime)
  \end{split}
\end{equation}
Where $P (\mathbf r, \mathbf r^\prime)$ is the object-independent system point-spread function (e.g. Eq. (\ref{eq:pinhole3})), describing the image produced by an isolated scatterer in the absence of any other scatterers or absorbers. 
We assume $P (\mathbf r, \mathbf r^\prime)$ is known,
it is a transfer function relating brightness at object points to signal at the detector.
$Q (\mathbf r, \mathbf r^\prime)$ is the object-dependent attenuation seen between a point in the object and a detection point.
\begin{equation} \label{eq:psf3}
  \begin{split}
    Q (\mathbf r, \mathbf r^\prime) = \text{exp} \left[-\int_\Omega \alpha (\mathbf r^{\prime \prime}) \delta^3 \left( r^{\prime \prime} \in L(\mathbf r^\prime, \mathbf r) \right) d \mathbf r^{\prime \prime} \right]
  \end{split}
\end{equation}
Eq. (\ref{eq:psf3}) describes the attenuation as the integral of $\alpha (\mathbf r^{\prime \prime})$, the attenuation coefficient, along the line segment $L(\mathbf r^\prime, \mathbf r)$, a ray path from $\mathbf r^{\prime}$  to $\mathbf r$, as in Eq. (\ref{generalray2}).

We define $\alpha(\mathbf r ) = \log \left( a( \mathbf r ) \right)$, 
and write Eq. (\ref{eq:green-incoherent2}) as
\begin{equation} \label{eq:green-incoherent4}
  \begin{split}
    I ( \mathbf r ) = \int_\Omega P (\mathbf r, \mathbf r^\prime) \; 
   \text{exp} \Bigl[-\int_\Omega \log \left( a( \mathbf r^{\prime \prime} ) \right) \delta^3 \left( \mathbf r^{\prime \prime} \in L(\mathbf r^\prime, \mathbf r) \right)  d \mathbf r^{\prime \prime} + \log \left( s( \mathbf r^\prime ) \right) \Bigr] d \mathbf r^\prime
  \end{split}
\end{equation}
$a (\mathbf r^\prime)$  and $b (\mathbf r^\prime)$ are the unknowns that we will be solving for, which describe the object in terms of the attenuation at each point and log of the brightness of each point, respectively. 

If $a (\mathbf r^\prime)$ is known, this means $Q (\mathbf r, \mathbf r^\prime)$ and therefore $H (\mathbf r, \mathbf r^\prime)$ is known, and we have a linear reconstruction problem to find $s( \mathbf r )$ using Eq. (\ref{eq:green-incoherent2}).
This is how Eq. (\ref{eq:green-incoherent4}) reduces to source-reconstruction tomography.
On the other hand, if $a (\mathbf r^\prime)$ is unknown but $s (\mathbf r^\prime)$ is known, we do not immediately get the logarithmic case of transmissive tomography.
We need the data collection designed such that at most a single source point is seen at each collection point, so that only a single ray is collected at each point on the detector.
Then the first integral in Eq. (\ref{eq:green-incoherent4}) is only over a single point, and we may take the log to get a linear system.

\subsection{Discrete domain} 

If we define the vector $\bf a$ as a sampled version of the function $a(\mathbf r)$, and 
$\bf s$ similarly from the function $s(\mathbf r)$, then our unknown is the vector $\bf x$  
formed by concatenating $\bf a$ and $\bf s$.
\begin{equation} \label{xab}
\mathbf x = \left(
  \begin{matrix}
    \mathbf a \\ 
    \mathbf s
  \end{matrix}
\right) = \left(
  \begin{matrix}
    a(\mathbf r_1) \\ 
  \vdots \\
    a(\mathbf r_K) \\ 
    s(\mathbf r_1) \\
  \vdots \\
    s(\mathbf r_K)
  \end{matrix}
\right)
\end{equation}
$K$ is the total number of pixels (or voxels) in the sampled image of the object.

The discrete form of Eq. (\ref{eq:green-incoherent2}), using discrete forms of $H(\mathbf r, \mathbf r^\prime)$, $P(\mathbf r, \mathbf r^\prime)$, and $Q(\mathbf r, \mathbf r^\prime)$,
is given by Eq. (\ref{eq:yPs}).
\begin{equation}
\label{eq:yPs} 
  \mathbf b = \mathbf H \, \mathbf s
            = ( \mathbf P \circ \mathbf Q ) \, \mathbf s
\end{equation}
Where $\bf b$ is the sampled data, and ``$\circ$'' is the Hadamard or element-wise product.
With this equation we can solve for cases with known attenuation by computing the resulting $\bb H$ matrix.

For unknown attenuation we write the discrete form of Eq. (\ref{eq:green-incoherent4}) as Eq. (\ref{eq:ycexpex}) where 
$\bf C$ is a matrix representing the integration over the kernel $P (\mathbf r, \mathbf r^\prime)$ followed by the Hadamard product, and $\bf E$ is a matrix representing the linear system inside the exponential (so both $\bf C$ and $\bf E$ are known).
\begin{equation}
\label{eq:ycexpex}
  \mathbf b = \mathbf C \; \exp \Bigl( \mathbf E \log \mathbf x \Bigr)
\end{equation}
The log and exponential are taken element-wise over vectors.

Note that we can write the equations of this nonlinear system as the multivariate system of polynomials,
\begin{equation}
\label{eq:poly}
    b_k = \sum_j C_{k,j} \; \prod_i x_i^{E_{j,i}} , \; k = 1 \cdots K 
\end{equation}
With the elements of $\mathbf C$ and $\mathbf E$ giving the coefficients and exponents, respectively.

\subsection{Solving via Optimization} 

Since Eq. (\ref{eq:ycexpex}) is nonlinear, we cannot easily use it as a constraint in a practical solution approach as we can with $\bb A \bb x = \bb b$.
Instead we form the mean-squared error between the measurements $\bf b$ and an estimate of the measurements using the forward model with variable $\bf x$,
\begin{align} \label{eq:lms_error}
    \phi(\mathbf x) &= \Vert \mathbf y - \mathbf C \, \textrm{exp} \{ \mathbf E \, \log \mathbf x \}\Vert_2^2 \notag \\
                    &= \mathbf r^T \mathbf r
\end{align}
The gradient of this expression is straightforward to compute, $\nabla \phi(\mathbf x) = 2 \mathbf r^T \mathbf J$. 
The Jacobian of the nonlinear system of Eq. (\ref{eq:ycexpex}) can be written as
\begin{align}
\label{eq:dydcexpex2_jacob}
  \mathbf J  = \mathbf C \; \text{diag} \bigl( \text{exp}  ( \mathbf E \log \mathbf x ) \bigr) \mathbf E
               \; \text{diag} \bigl( \mathbf x^{(-1)} \bigr)
\end{align}
Where $\text{diag}(\mathbf v)$ is the diagonal matrix with diagonal $\bf v$ and $\mathbf x^{(-1)}$ denotes the vector with elements $(\mathbf x^{(-1)})_i = (x_i)^{-1}$.


Next we consider the appropriate way to regularize this system.
Recall that our unknown vector $\bf x$ consists of two distinct components, independently describing a brightness and an attenuation. 
The choice of what is the more ``regular'' $\bf x$, therefore, must consider each component separately.
We will regularize by choosing the most probably solution to be empty space for each point.
The values of our variables for empty space are given in Table \ref{airtable}, which lists the extremes of empty space and opaque matter, as well as the total range.
\begin{table}
\begin{center}
\begin{tabular}{ r | c | c | c } 
              & Air   & Solid & Range \\ \hline
  Attenuation & $a=1$ & $a=0$ & $1 \ge a \ge 0$  \\
  Brightness  & $s=0$ & $s>0$ & $s \ge 0$        \\
\end{tabular}
\end{center}
\caption{Unknowns for points in empty space (``air''), in opaque solid, and the net range.}
\label{airtable}
\end{table}
Using Eq. (\ref{xab}), this means if our sample consisted of nothing but empty space, $\bf x$ would be 
\begin{equation} \label{xair}
\mathbf x_{air} = \left(
  \begin{matrix}
    \mathbf 1 \\ 
    \mathbf 0
  \end{matrix}
\right) = \left(
  \begin{matrix}
    1 \\ 
  \vdots \\
    1 \\ 
    0 \\
  \vdots \\
    0
  \end{matrix}
\right)
\end{equation} 
We also see from Table \ref{airtable} that we have the constraint, $\bf x \ge 0$, but to avoid problems with the log at zero, we will set the minimum to be some small value $\epsilon > 0$, much smaller than noticeable object values.
For example we use $\epsilon = 10^{-9}$ with brightness values on the order of $10^0$.
Putting everything together, we have as the optimization problem, Eq. (\ref{eq:121911-2}).
\begin{equation} \label{eq:121911-2}
  \begin{array}{c}
    \underset{\mathbf x}{\textrm{min}}  \Vert \mathbf y - \mathbf C \, \textrm{exp} \{ \mathbf E \log \, \mathbf x \}\Vert_2^2  + \mu \Vert \mathbf{x} - \mathbf x_{air} \Vert_2^2\\
    \mathbf{ x \ge \epsilon }
  \end{array}
\end{equation}
The scalar constant $\mu$ is the regularization parameter, which we choose to be higher than the noise level.

\begin{figure}[h] \centering 
\begin{tikzpicture}[font=\small]
  \begin{axis}[%
  width=1.5in,
  height=1.5in,
  axis on top,
  clip=false,
  scale only axis,
  xmin=0.5,
  xmax=50.5,
  xtick={ 1, 10, 20, 30, 40, 50},
  y dir=reverse,
  ymin=0.5,
  ymax=50.5,
  ytick={ 50, 40, 30, 20, 10, 1},
  title={True $\bb a$},
  colormap/hot2,
  colorbar,
  colorbar style={ytick={  0, 0.2, 0.4, 0.6, 0.8,   1}},
  point meta min=3.72007597602084e-44,
  point meta max=1
  ]
  \addplot graphics [xmin=0.5,xmax=50.5,ymin=0.5,ymax=50.5] {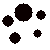};
  \node[right, inner sep=0mm, text=black]
  at (axis cs:-2,-2,0) {(a)};
  \end{axis}
\end{tikzpicture}%
\begin{tikzpicture}[font=\small]
  \begin{axis}[%
  width=1.5in,
  height=1.5in,
  axis on top,
  clip=false,
  scale only axis,
  xmin=0.5,
  xmax=50.5,
  xtick={ 1, 10, 20, 30, 40, 50},
  y dir=reverse,
  ymin=0.5,
  ymax=50.5,
  ytick={ 50, 40, 30, 20, 10, 1},
  title={True $\bb s$},
  colormap={mymap}{[1pt] rgb(0pt)=(1,1,1); rgb(16pt)=(1,1,0); rgb(40pt)=(1,0,0); rgb(63pt)=(0.0416667,0,0)},
  colorbar,
  point meta min=5.48776088326244e-44,
  point meta max=1
  ]
  \addplot graphics [xmin=0.5,xmax=50.5,ymin=0.5,ymax=50.5] {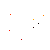};
  \node[right, inner sep=0mm, text=black]
  at (axis cs:-2,-2,0) {(b)};
  \end{axis}
\end{tikzpicture}%
\caption{True versus estimated attenuation and brightness images for object consisting of dark opaque objects, and seven point sources.} 
\label{fig:qdot6}
\end{figure}

To demonstrate the solution of independent attenuation and brightness maps in the presence of occlusion, we simulated an object with an attenuation map consisting of five opaque circles, and a brightness map consisting of seven point sources.
We used a collection geometry similar to computed tomography: views were collected at each angle in one-degree increments in a complete circle around the object.
The simulated object is shown in Fig. \ref{fig:qdot6}, where the $\bf a$ and $\bf s$ components of the true $\bf x$ vector are arranged into images.
The attenuation map is given in Fig. \ref{fig:qdot6}(a).
Inside the objects, $a = 0$, so the objects are completely opaque, whereas the air has $a=1$, so there is no attenuation.  
Note that the brightness map of Fig. \ref{fig:qdot6}(b) has a reversed colormap to set the color of air to white, as air is zero brightness but unit attenuation.

\begin{figure}[h] \centering 
\begin{tikzpicture}[font=\small]
\begin{axis}[%
width=5.1in,
height=1.5in,
axis on top,
scale only axis,
xmin=0.5,
xmax=360.5,
xtick={  1,  50, 100, 150, 200, 250, 300, 350},
xlabel={view direction (degrees)},
y dir=reverse,
ymin=0.5,
ymax=50.5,
ytick={ 1, 10, 20, 30, 40, 50},
ylabel={projection image pixel},
title={Opaque slant stack},
colormap={mymap}{[1pt] rgb(0pt)=(1,1,1); rgb(16pt)=(1,1,0); rgb(40pt)=(1,0,0); rgb(63pt)=(0.0416667,0,0)},
colorbar,
colorbar style={ytick={  0, 0.2, 0.4, 0.6, 0.8,   1}},
point meta min=0,
point meta max=1
]
\addplot graphics [xmin=0.5,xmax=360.5,ymin=0.5,ymax=50.5] {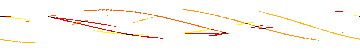};
\end{axis}
\end{tikzpicture}
\caption{Slant stack for seven-source object with five large occluders.}
\label{qdot7_opaq}
\end{figure}

\begin{figure}[h] \centering 
\begin{tikzpicture}[font=\small]
\begin{axis}[%
width=5.1in,
height=1.5in,
axis on top,
scale only axis,
xmin=0.5,
xmax=360.5,
xtick={  1,  50, 100, 150, 200, 250, 300, 350},
xlabel={view direction (degrees)},
y dir=reverse,
ymin=0.5,
ymax=50.5,
ytick={ 1, 10, 20, 30, 40, 50},
ylabel={projection image pixel},
title={Linear slant stack},
colormap={mymap}{[1pt] rgb(0pt)=(1,1,1); rgb(16pt)=(1,1,0); rgb(40pt)=(1,0,0); rgb(63pt)=(0.0416667,0,0)},
colorbar,
colorbar style={ytick={  0, 0.2, 0.4, 0.6, 0.8,   1}},
point meta min=0,
point meta max=1
]
\addplot graphics [xmin=0.5,xmax=360.5,ymin=0.5,ymax=50.5] {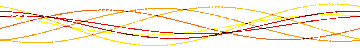};
\end{axis}
\end{tikzpicture}%
\caption{Slant stack for seven-source object neglecting occluders.}
\label{qdot7_y}
\end{figure}

The slant stack for a full 360-degree collection is shown in Fig. \ref{qdot7_opaq}.
As the object is two-dimensional, each ``view'' is one-dimensional (edgewise), given as the columns in the image in Fig. \ref{qdot7_opaq}.
The horizontal axis gives the angle at which the view was collected. 
This is a generalization of the definition of the slant stack, as the values at points are no longer linear projections along rays, but attenuated projections of sources along the rays.
Note that the sources are only visible at certain intervals, when they are not occluded. 
Fig. \ref{qdot7_y} shows what the slant stack would look like in the absense of any attenuation (i.e. a linear system with simple projections along the rays). 

A separate algorithm was used to generate the occluded data, based on determining the brightness of the first-seen opaque point for each detector point.
Also noise was added to the data at roughly one percent of the signal level.
To solve the optimization problem, we used SNOPT \cite{gill_snopt:_2005}, a nonlinear optimization solver.

\begin{figure}[h] \centering 
\begin{tikzpicture}[font=\small]
  \begin{axis}[%
  width=1.5in,
  height=1.5in,
  axis on top,
  clip=false,
  scale only axis,
  xmin=0.5,
  xmax=50.5,
  xtick={ 1, 10, 20, 30, 40, 50},
  y dir=reverse,
  ymin=0.5,
  ymax=50.5,
  ytick={ 50, 40, 30, 20, 10, 1},
  title={Estimated $\bb a$},
  colormap/hot2,
  colorbar,
  colorbar style={ytick={  0, 0.2, 0.4, 0.6, 0.8,   1}},
  point meta min=0,
  point meta max=1
  ]
  \addplot graphics [xmin=0.5,xmax=50.5,ymin=0.5,ymax=50.5] {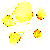};
  \node[right, inner sep=0mm, text=black]
  at (axis cs:-2,-2,0) {(a)};
  \end{axis}
\end{tikzpicture}%
\begin{tikzpicture}[font=\small]
  \begin{axis}[%
  width=1.5in,
  height=1.5in,
  axis on top,
  clip=false,
  scale only axis,
  xmin=0.5,
  xmax=50.5,
  xtick={ 1, 10, 20, 30, 40, 50},
  y dir=reverse,
  ymin=0.5,
  ymax=50.5,
  ytick={ 50, 40, 30, 20, 10, 1},
  title={Estimated $\bb s$},
  colormap={mymap}{[1pt] rgb(0pt)=(1,1,1); rgb(16pt)=(1,1,0); rgb(40pt)=(1,0,0); rgb(63pt)=(0.0416667,0,0)},
  colorbar,
  point meta min=0,
  point meta max=1
  ]
  \addplot graphics [xmin=0.5,xmax=50.5,ymin=0.5,ymax=50.5] {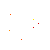};
  \node[right, inner sep=0mm, text=black]
  at (axis cs:-2,-2,0) {(b)};
  \end{axis}
\end{tikzpicture}%
\caption{Estimated attenuation (a) and brightness (b) images for object consisting of dark opaque objects, and seven point sources.} 
\label{fig:qdot6_est}
\end{figure}

\begin{figure}[h] \centering 
\begin{tikzpicture}[font=\small]
\begin{axis}[%
width=1.5in,
height=1.5in,
axis on top,
clip=false,
scale only axis,
xmin=0.5,
xmax=50.5,
xtick={ 1, 10, 20, 30, 40, 50},
y dir=reverse,
ymin=0.5,
ymax=50.5,
ytick={ 1, 10, 20, 30, 40, 50},
title={True $\bb a$},
colormap/hot2,
colorbar,
point meta min=3.72007597602084e-44,
point meta max=1
]
\addplot graphics [xmin=0.5,xmax=50.5,ymin=0.5,ymax=50.5] {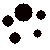};
\node[right, inner sep=0mm, text=black]
at (axis cs:-2,-2,0) {(a)};
\end{axis}
\end{tikzpicture}%
\begin{tikzpicture}[font=\small]
\begin{axis}[%
width=1.5in,
height=1.5in,
axis on top,
clip=false,
scale only axis,
xmin=0.5,
xmax=50.5,
xtick={ 1, 10, 20, 30, 40, 50},
y dir=reverse,
ymin=0.5,
ymax=50.5,
ytick={ 1, 10, 20, 30, 40, 50},
title={True $\bb s$},
colormap={mymap}{[1pt] rgb(0pt)=(1,1,1); rgb(16pt)=(1,1,0); rgb(40pt)=(1,0,0); rgb(63pt)=(0.0416667,0,0)},
colorbar,
point meta min=5.11036006472054e-44,
point meta max=1
]
\addplot graphics [xmin=0.5,xmax=50.5,ymin=0.5,ymax=50.5] {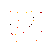};
\node[right, inner sep=0mm, text=black]
at (axis cs:-2,-2,0) {(b)};
\end{axis}
\end{tikzpicture}%
\\
\begin{tikzpicture}[font=\small]
\begin{axis}[%
width=1.5in,
height=1.5in,
axis on top,
clip=false,
scale only axis,
xmin=0.5,
xmax=50.5,
xtick={ 1, 10, 20, 30, 40, 50},
y dir=reverse,
ymin=0.5,
ymax=50.5,
ytick={ 1, 10, 20, 30, 40, 50},
title={Estimated $\bb a$},
colormap/hot2,
colorbar,
point meta min=0,
point meta max=1
]
\addplot graphics [xmin=0.5,xmax=50.5,ymin=0.5,ymax=50.5] {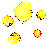};
\node[right, inner sep=0mm, text=black]
at (axis cs:-2,-2,0) {(c)};
\end{axis}
\end{tikzpicture}%
\begin{tikzpicture}[font=\small]
\begin{axis}[%
width=1.5in,
height=1.5in,
axis on top,
clip=false,
scale only axis,
xmin=0.5,
xmax=50.5,
xtick={ 1, 10, 20, 30, 40, 50},
y dir=reverse,
ymin=0.5,
ymax=50.5,
ytick={ 1, 10, 20, 30, 40, 50},
title={Estimated $\bb s$},
colormap={mymap}{[1pt] rgb(0pt)=(1,1,1); rgb(16pt)=(1,1,0); rgb(40pt)=(1,0,0); rgb(63pt)=(0.0416667,0,0)},
colorbar,
point meta min=0,
point meta max=1
]
\addplot graphics [xmin=0.5,xmax=50.5,ymin=0.5,ymax=50.5] {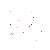};
\node[right, inner sep=0mm, text=black]
at (axis cs:-2,-2,0) {(d)};
\end{axis}
\end{tikzpicture}%
\caption{True versus estimated attenuation and brightness images for object consisting of dark occluders and 20 point sources.} 
\label{fig:qdot20}
\end{figure}
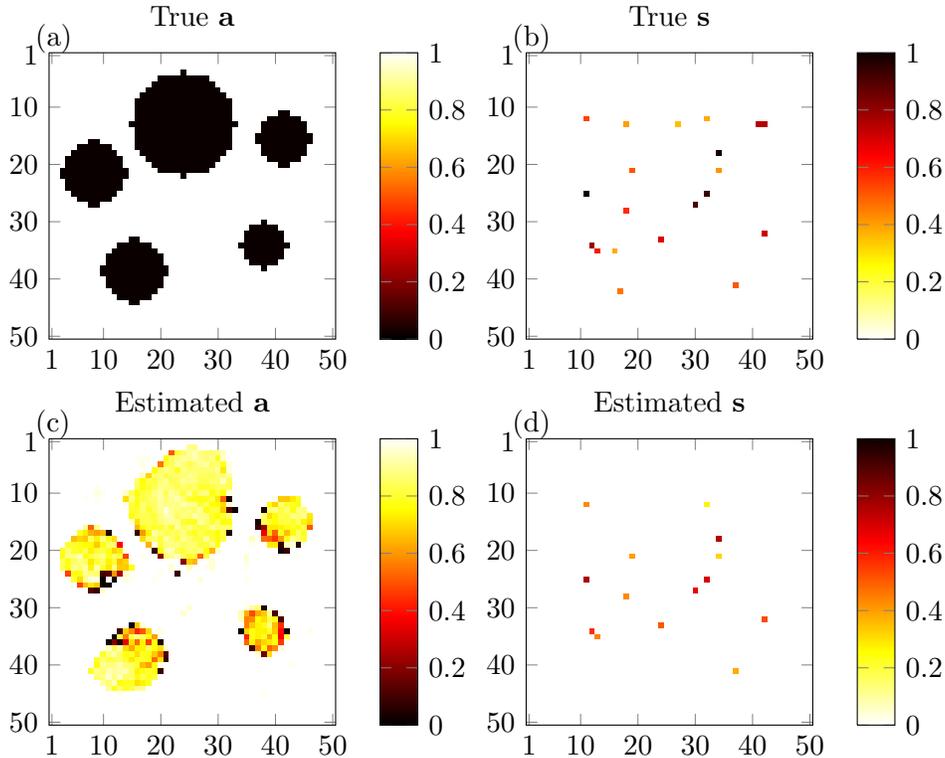

The reconstructed results (using the slant stack of Fig. \ref{qdot7_opaq}) are given in Figs. \ref{fig:qdot6_est}. 
We find we have an easy time imaging the point sources themselves (one of them was not reconstructed as it was inside an opaque region), suggesting an advantage gained from sparsity.
In Fig. \ref{fig:qdot20} we give the simulated attenuation and brightness and estimated results using 20 point sources, and note a significant improvement from the additional sources.

\subsection{Multiview Reconstruction}



\begin{figure}[h] \centering 
\begin{tikzpicture}[font=\small]
\begin{axis}[%
width=1.5in,
height=1.5in,
axis on top,
clip=false,
scale only axis,
xmin=0.5,
xmax=50.5,
xtick={ 1, 10, 20, 30, 40, 50},
y dir=reverse,
ymin=0.5,
ymax=50.5,
ytick={ 1, 10, 20, 30, 40, 50},
title={True $\bb a$},
colormap/hot2,
colorbar,
point meta min=3.72007597602084e-44,
point meta max=1
]
\addplot graphics [xmin=0.5,xmax=50.5,ymin=0.5,ymax=50.5] {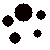};
\node[right, inner sep=0mm, text=black]
at (axis cs:-2,-2,0) {(a)};
\end{axis}
\end{tikzpicture}%
\begin{tikzpicture}[font=\small]
\begin{axis}[%
width=1.5in,
height=1.5in,
axis on top,
clip=false,
scale only axis,
xmin=0.5,
xmax=50.5,
xtick={ 1, 10, 20, 30, 40, 50},
y dir=reverse,
ymin=0.5,
ymax=50.5,
ytick={ 1, 10, 20, 30, 40, 50},
title={True $\bb s$},
colormap={mymap}{[1pt] rgb(0pt)=(1,1,1); rgb(16pt)=(1,1,0); rgb(40pt)=(1,0,0); rgb(63pt)=(0.0416667,0,0)},
colorbar,
point meta min=4.13341775113426e-44,
point meta max=1
]
\addplot graphics [xmin=0.5,xmax=50.5,ymin=0.5,ymax=50.5] {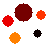};
\node[right, inner sep=0mm, text=black]
at (axis cs:-2,-2,0) {(b)};
\end{axis}
\end{tikzpicture}%
\caption{Attenuation (a) and brightness (b) maps for opaque object consisting of five circles. Note the colormaps are inverses so that free space is consistently white.} 
\label{fig:Multiview Reconstruction}
\end{figure}

Next we made a two-dimensional simulation of an opaque object which scattered light as the brightness source itself. 
The simulated object is shown in Fig. \ref{fig:Multiview Reconstruction}. 
Note that the brightness is nonzero whenever the attenuation is zero, denoting opaque matter. 
The brightness on the interior of the opaque regions is meaningless, it cannot be seen due to occlusion by the opaque points on the exterior of the opaque regions.

\begin{figure}[h] \centering 
\begin{tikzpicture}[font=\small]
\begin{axis}[%
width=5.1in,
height=1.5in,
axis on top,
scale only axis,
xmin=0.5,
xmax=360.5,
xtick={  1,  50, 100, 150, 200, 250, 300, 350},
xlabel={view direction (degrees)},
y dir=reverse,
ymin=0.5,
ymax=50.5,
ytick={ 1, 10, 20, 30, 40, 50},
ylabel={projection image pixel},
title={Opaque slant stack},
colormap={mymap}{[1pt] rgb(0pt)=(1,1,1); rgb(16pt)=(1,1,0); rgb(40pt)=(1,0,0); rgb(63pt)=(0.0416667,0,0)},
colorbar,
point meta min=0,
point meta max=1
]
\addplot graphics [xmin=0.5,xmax=360.5,ymin=0.5,ymax=50.5] {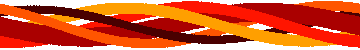};
\end{axis}
\end{tikzpicture}%
\caption{Data from multiple views of opaque system of Fig. \ref{fig:Multiview Reconstruction}. Each column represents an image for one view edgewise of the brightness map, subject to the attenuation map. }
\label{fig:nonlineartomoydata}
\end{figure}

\begin{figure}[h] \centering 
\begin{tikzpicture}[font=\small]
\begin{axis}[%
width=5.1in,
height=1.5in,
axis on top,
scale only axis,
xmin=0.5,
xmax=360.5,
xtick={  1,  50, 100, 150, 200, 250, 300, 350},
xlabel={view direction (degrees)},
y dir=reverse,
ymin=0.5,
ymax=50.5,
ytick={ 1, 10, 20, 30, 40, 50},
ylabel={projection image pixel},
title={Linear slant stack},
colormap={mymap}{[1pt] rgb(0pt)=(1,1,1); rgb(16pt)=(1,1,0); rgb(40pt)=(1,0,0); rgb(63pt)=(0.0416667,0,0)},
colorbar,
colorbar style={ytick={ 0,  5, 10, 15, 20, 25}},
point meta min=0,
point meta max=25
]
\addplot graphics [xmin=0.5,xmax=360.5,ymin=0.5,ymax=50.5] {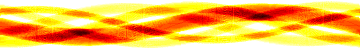};
\end{axis}
\end{tikzpicture}%
\caption{Slant stack of projections of brightness map alone, attenuation is neglected.} 
\label{multiview_ylin}
\end{figure}

\begin{figure}[h] \centering 
\begin{tikzpicture}[font=\small]
\begin{axis}[%
width=1.5in,
height=1.5in,
axis on top,
clip=false,
scale only axis,
xmin=0.5,
xmax=50.5,
xtick={ 1, 10, 20, 30, 40, 50},
y dir=reverse,
ymin=0.5,
ymax=50.5,
ytick={ 1, 10, 20, 30, 40, 50},
title={Estimated $\bb a$},
colormap/hot2,
colorbar,
point meta min=0,
point meta max=1
]
\addplot graphics [xmin=0.5,xmax=50.5,ymin=0.5,ymax=50.5] {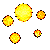};
\node[right, inner sep=0mm, text=black]
at (axis cs:-2,-2,0) {(a)};
\end{axis}
\end{tikzpicture}%
\begin{tikzpicture}[font=\small]
\begin{axis}[%
width=1.5in,
height=1.5in,
axis on top,
clip=false,
scale only axis,
xmin=0.5,
xmax=50.5,
xtick={ 1, 10, 20, 30, 40, 50},
y dir=reverse,
ymin=0.5,
ymax=50.5,
ytick={ 1, 10, 20, 30, 40, 50},
title={Estimated $\bb s$},
colormap={mymap}{[1pt] rgb(0pt)=(1,1,1); rgb(16pt)=(1,1,0); rgb(40pt)=(1,0,0); rgb(63pt)=(0.0416667,0,0)},
colorbar,
point meta min=0,
point meta max=1
]
\addplot graphics [xmin=0.5,xmax=50.5,ymin=0.5,ymax=50.5] {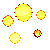};
\node[right, inner sep=0mm, text=black]
at (axis cs:-2,-2,0) {(b)};
\end{axis}
\end{tikzpicture}%

\caption{Reconstructed attenuation (a) and brightness (b) components of the variable $\bf x$.}
\label{fig:Multiviewx}
\end{figure}

The simulated slant stack $\bf b$ is given in Fig. \ref{fig:nonlineartomoydata}. 
We can see how the different components occlude each other from different directions.
Again we give the linear version of the slant stack neglecting occlusion in Fig. \ref{multiview_ylin}.
In Fig. \ref{fig:Multiviewx}, the reconstruction results are given when using the full 360 degree data collection from Fig. \ref{fig:Multiview Reconstruction}.
The result is a relatively accurate estimate of the truth (Fig. \ref{fig:Multiview Reconstruction}); the interiors of the opaque objects are of course missing as they are completely hidden. 
Note that the brightness estimate of the surfaces has the correct brightness level.

\section{Exploiting Self-Occlusion}

Thus far we have seen self-occlusion within the object as a purely negative effect.
Full ``slant stacks'' of data that would suffice for reconstruction in the linear case, have information missing due to occlusion.
But now we consider that occlusion may be utilized to provide diversity in the collected data and help in reconstruction.

\begin{figure}[h] \centering 
\input{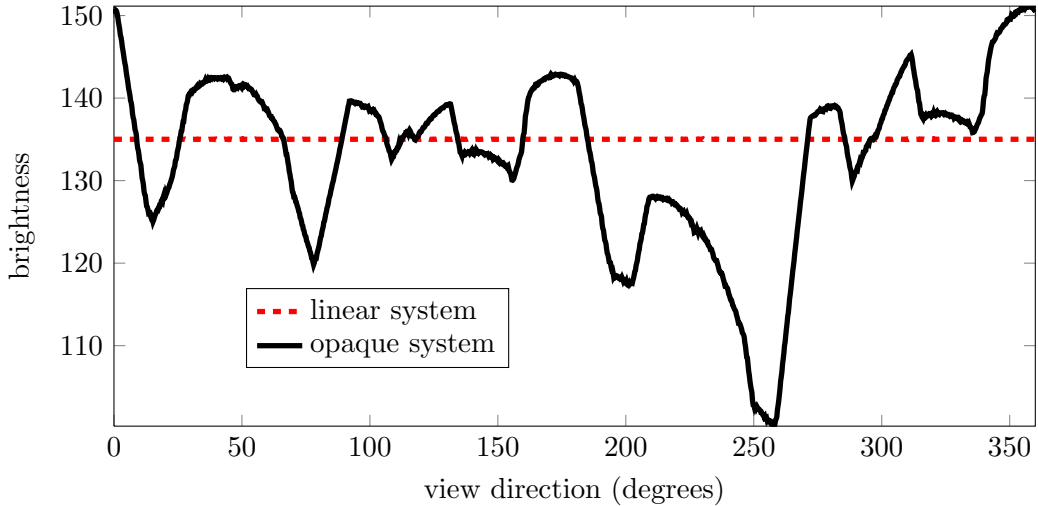}
\caption{Linear versus opaque lightcurves for object consisting of five opaque regions.} 
\label{lin_vs_opaq_lc}
\end{figure}

We would expect that occlusion might help super-resolve low resolution data, since the occlusion occurs at a high resolution within the object.
In particular we demonstrate the extreme case where we only collect a single brightness value per view, effectively the sum over the vertical dimension of the slant stack.
We can formulate this as the sum over views in the measured data of Eq. (\ref{eq:ycexpex}).
For example for a single view we would get
\begin{align}
\label{eq:dydcexpex}
  b_{lc} & =  \sum_i^n (\mathbf y)_i \notag \\
         & =  \mathbf 1^T \mathbf y \notag \\
         & = \mathbf 1^T \mathbf C \; \text{exp} \Bigl( \mathbf E \mathbf x \Bigr)
\end{align}
$b_{lc}$ is the brightness at a single view.
Over a range of views this would form a so-called lightcurve and we would have a system of the form of Eq. (\ref{eq:dydcexpex_2}).
\begin{align}
\label{eq:dydcexpex_2}
  \bb b_{lc} & = \mathbf D \mathbf y \notag \\
             & = \mathbf D \mathbf C \; \text{exp} \Bigl( \mathbf E \mathbf x \Bigr)
\end{align}
Where $\bb D$ is a matrix that performs the downsampling to one measurement per view.
An example is given in Fig. \ref{lin_vs_opaq_lc}, where we show the result for the slant stack of Fig. \ref{fig:nonlineartomoydata}
versus the lightcurve that results from the linear case of Fig. \ref{multiview_ylin}.
Here we can clearly see the gain in signal diversity that occlusion provides.

\subsection{Lightcurve Imaging}

In the lightcurve inversion problem \cite{wild_light-curve_1991, kaasalainen_optimization_2001-1, kaasalainen_optimization_2001, masiero_thousand_2009}, a distant object such as a planet or asteroid is too far to resolve with measurements in a single aperture.
Only a single point of brightness can be seen.
The so-called lightcurve is the trace of this brightness level over time.
If we presume these variations are solely the result of the object's rotation causing differences in views of the object presented, then we might use the lightcurve to estimate the shape of the object.
The problem is also referred to as photometric signature inversion \cite{calef_photometric_2006}, and 
reconstruction from brightness functions \cite{gardner_reconstruction_2003}. 
Generally this is solved using a parametric approach that assumes the object has the shape of a convex polytope with predetermined number of sides.

\begin{figure}[h] \centering 
\input{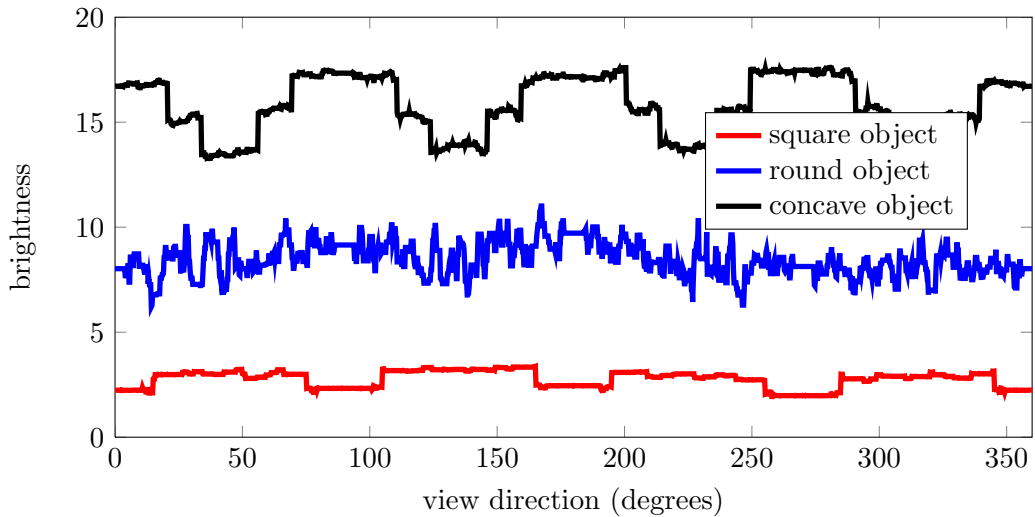}
\caption{Lightcurves for the three simulated objects.} 
\label{lc_y}
\end{figure}

Now we will simulate a non-parametric reconstruction of this case in two dimensions. 
Give the extreme shortage of information, even in the two-dimensional case, we must restrict the problem to much simpler objects. 
Simulations were performed using a square object, a circular object, and a cross-shaped object.
The lightcurves for these objects are plotted in Fig. \ref{lc_y}

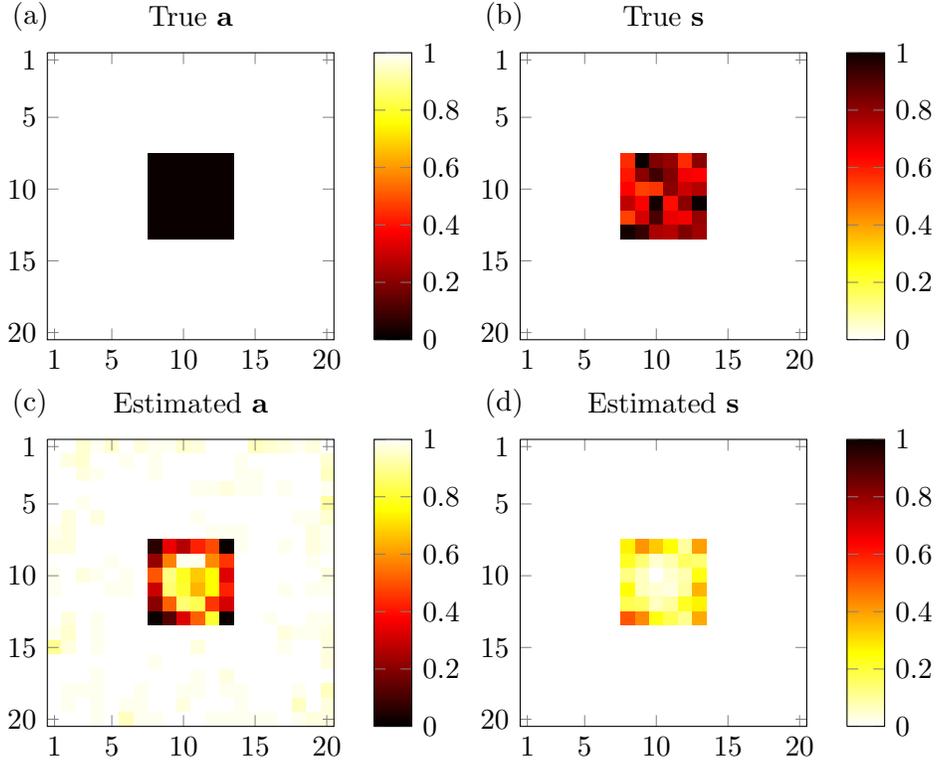
\begin{figure}[h] \centering 
\begin{tikzpicture}[font=\small]
\begin{axis}[%
width=1.5in,
height=1.50333333333333in,
axis on top,
clip=false,
scale only axis,
xmin=0.5,
xmax=20.5,
xtick={ 1,  5, 10, 15, 20},
y dir=reverse,
ymin=0.5,
ymax=20.5,
ytick={ 1,  5, 10, 15, 20},
title={True $\bb a$},
colormap/hot2,
colorbar,
colorbar style={ytick={  0, 0.2, 0.4, 0.6, 0.8,   1}},
point meta min=3.72007597602084e-44,
point meta max=1
]
\addplot graphics [xmin=0.5,xmax=20.5,ymin=0.5,ymax=20.5] {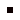};
\node[right, inner sep=0mm, text=black]
at (axis cs:-2,-2,0) {(a)};
\end{axis}
\end{tikzpicture}%
\begin{tikzpicture}[font=\small]

\begin{axis}[%
width=1.5in,
height=1.50333333333333in,
axis on top,
clip=false,
scale only axis,
xmin=0.5,
xmax=20.5,
xtick={ 1,  5, 10, 15, 20},
y dir=reverse,
ymin=0.5,
ymax=20.5,
ytick={ 1,  5, 10, 15, 20},
title={True $\bb s$},
colormap={mymap}{[1pt] rgb(0pt)=(1,1,1); rgb(16pt)=(1,1,0); rgb(40pt)=(1,0,0); rgb(63pt)=(0.0416667,0,0)},
colorbar,
point meta min=7.5800472603689e-44,
point meta max=1
]
\addplot graphics [xmin=0.5,xmax=20.5,ymin=0.5,ymax=20.5] {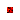};
\node[right, inner sep=0mm, text=black]
at (axis cs:-2,-2,0) {(b)};
\end{axis}
\end{tikzpicture}%
\\
\begin{tikzpicture}[font=\small]

\begin{axis}[%
width=1.5in,
height=1.50333333333333in,
axis on top,
clip=false,
scale only axis,
xmin=0.5,
xmax=20.5,
xtick={ 1,  5, 10, 15, 20},
y dir=reverse,
ymin=0.5,
ymax=20.5,
ytick={ 1,  5, 10, 15, 20},
title={Estimated $\bb a$},
colormap/hot2,
colorbar,
point meta min=0,
point meta max=1
]
\addplot graphics [xmin=0.5,xmax=20.5,ymin=0.5,ymax=20.5] {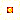};
\node[right, inner sep=0mm, text=black]
at (axis cs:-2,-2,0) {(c)};
\end{axis}
\end{tikzpicture}%
\begin{tikzpicture}[font=\small]

\begin{axis}[%
width=1.5in,
height=1.50333333333333in,
axis on top,
clip=false,
scale only axis,
xmin=0.5,
xmax=20.5,
xtick={ 1,  5, 10, 15, 20},
y dir=reverse,
ymin=0.5,
ymax=20.5,
ytick={ 1,  5, 10, 15, 20},
title={Estimated $\bb s$},
colormap={mymap}{[1pt] rgb(0pt)=(1,1,1); rgb(16pt)=(1,1,0); rgb(40pt)=(1,0,0); rgb(63pt)=(0.0416667,0,0)},
colorbar,
point meta min=0,
point meta max=1
]
\addplot graphics [xmin=0.5,xmax=20.5,ymin=0.5,ymax=20.5] {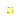};
\node[right, inner sep=0mm, text=black]
at (axis cs:-2,-2,0) {(d)};
\end{axis}
\end{tikzpicture}%
\caption{True attenuation (a) and brightness (b) components of the variable $\bf x$.
Reconstructed attenuation (c) and brightness (d) components.}
\label{lcsquare}
\end{figure}

\begin{figure}[h] \centering 
\begin{tikzpicture}[font=\small]

\begin{axis}[%
width=1.5in,
height=1.50333333333333in,
axis on top,
clip=false,
scale only axis,
xmin=0.5,
xmax=32.5,
xtick={ 1,  5, 10, 15, 20, 25, 30},
y dir=reverse,
ymin=0.5,
ymax=32.5,
ytick={ 1,  5, 10, 15, 20, 25, 30},
title={True $\bb a$},
colormap/hot2,
colorbar,
colorbar style={ytick={  0, 0.2, 0.4, 0.6, 0.8,   1}},
point meta min=3.72007597602084e-44,
point meta max=1
]
\addplot graphics [xmin=0.5,xmax=32.5,ymin=0.5,ymax=32.5] {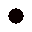};
\node[right, inner sep=0mm, text=black]
at (axis cs:-2,-2,0) {(a)};
\end{axis}
\end{tikzpicture}%
\begin{tikzpicture}[font=\small]

\begin{axis}[%
width=1.5in,
height=1.50333333333333in,
axis on top,
clip=false,
scale only axis,
xmin=0.5,
xmax=32.5,
xtick={ 1,  5, 10, 15, 20, 25, 30},
y dir=reverse,
ymin=0.5,
ymax=32.5,
ytick={ 1,  5, 10, 15, 20, 25, 30},
title={True $\bb s$},
colormap={mymap}{[1pt] rgb(0pt)=(1,1,1); rgb(16pt)=(1,1,0); rgb(40pt)=(1,0,0); rgb(63pt)=(0.0416667,0,0)},
colorbar,
colorbar style={ytick={  0, 0.2, 0.4, 0.6, 0.8,   1}},
point meta min=3.73730597439907e-44,
point meta max=1
]
\addplot graphics [xmin=0.5,xmax=32.5,ymin=0.5,ymax=32.5] {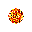};
\node[right, inner sep=0mm, text=black]
at (axis cs:-2,-2,0) {(b)};
\end{axis}
\end{tikzpicture}%
\\
\begin{tikzpicture}[font=\small]

\begin{axis}[%
width=1.5in,
height=1.50333333333333in,
axis on top,
clip=false,
scale only axis,
xmin=0.5,
xmax=32.5,
xtick={ 1,  5, 10, 15, 20, 25, 30},
y dir=reverse,
ymin=0.5,
ymax=32.5,
ytick={ 1,  5, 10, 15, 20, 25, 30},
title={Estimated $\bb a$},
colormap/hot2,
colorbar,
colorbar style={ytick={  0, 0.2, 0.4, 0.6, 0.8,   1}},
point meta min=0,
point meta max=1
]
\addplot graphics [xmin=0.5,xmax=32.5,ymin=0.5,ymax=32.5] {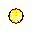};
\node[right, inner sep=0mm, text=black]
at (axis cs:-2,-2,0) {(c)};
\end{axis}
\end{tikzpicture}%
\begin{tikzpicture}[font=\small]

\begin{axis}[%
width=1.5in,
height=1.50333333333333in,
axis on top,
clip=false,
scale only axis,
xmin=0.5,
xmax=32.5,
xtick={ 1,  5, 10, 15, 20, 25, 30},
y dir=reverse,
ymin=0.5,
ymax=32.5,
ytick={ 1,  5, 10, 15, 20, 25, 30},
title={Estimated $\bb s$},
colormap={mymap}{[1pt] rgb(0pt)=(1,1,1); rgb(16pt)=(1,1,0); rgb(40pt)=(1,0,0); rgb(63pt)=(0.0416667,0,0)},
colorbar,
colorbar style={ytick={  0, 0.2, 0.4, 0.6, 0.8,   1}},
point meta min=0,
point meta max=1
]
\addplot graphics [xmin=0.5,xmax=32.5,ymin=0.5,ymax=32.5] {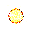};
\node[right, inner sep=0mm, text=black]
at (axis cs:-2,-2,0) {(d)};
\end{axis}
\end{tikzpicture}%
\caption{True attenuation (a) and brightness (b) components of the variable $\bf x$.
Reconstructed attenuation (c) and brightness (d) components.}
\label{lccirc}
\end{figure}

\begin{figure}[h] \centering 
\begin{tikzpicture}[font=\small]

\begin{axis}[%
width=1.5in,
height=1.50333333333333in,
axis on top,
clip=false,
scale only axis,
xmin=0.5,
xmax=32.5,
xtick={ 1,  5, 10, 15, 20, 25, 30},
y dir=reverse,
ymin=0.5,
ymax=32.5,
ytick={ 1,  5, 10, 15, 20, 25, 30},
title={True $\bb a$},
colormap/hot2,
colorbar,
colorbar style={ytick={  0, 0.2, 0.4, 0.6, 0.8,   1}},
point meta min=3.72007597602084e-44,
point meta max=1
]
\addplot graphics [xmin=0.5,xmax=32.5,ymin=0.5,ymax=32.5] {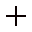};
\node[right, inner sep=0mm, text=black]
at (axis cs:-2,-2,0) {(a)};
\end{axis}
\end{tikzpicture}%
\begin{tikzpicture}[font=\small]

\begin{axis}[%
width=1.5in,
height=1.50333333333333in,
axis on top,
clip=false,
scale only axis,
xmin=0.5,
xmax=32.5,
xtick={ 1,  5, 10, 15, 20, 25, 30},
y dir=reverse,
ymin=0.5,
ymax=32.5,
ytick={ 1,  5, 10, 15, 20, 25, 30},
title={True $\bb s$},
colormap={mymap}{[1pt] rgb(0pt)=(1,1,1); rgb(16pt)=(1,1,0); rgb(40pt)=(1,0,0); rgb(63pt)=(0.0416667,0,0)},
colorbar,
colorbar style={ytick={  0, 0.2, 0.4, 0.6, 0.8,   1}},
point meta min=3.72436856436948e-44,
point meta max=1
]
\addplot graphics [xmin=0.5,xmax=32.5,ymin=0.5,ymax=32.5] {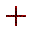};
\node[right, inner sep=0mm, text=black]
at (axis cs:-2,-2,0) {(b)};
\end{axis}
\end{tikzpicture}%
\\
\begin{tikzpicture}[font=\small]

\begin{axis}[%
width=1.5in,
height=1.50333333333333in,
axis on top,
clip=false,
scale only axis,
xmin=0.5,
xmax=32.5,
xtick={ 1,  5, 10, 15, 20, 25, 30},
y dir=reverse,
ymin=0.5,
ymax=32.5,
ytick={ 1,  5, 10, 15, 20, 25, 30},
title={Estimated $\bb a$},
colormap/hot2,
colorbar,
colorbar style={ytick={  0, 0.2, 0.4, 0.6, 0.8,   1}},
point meta min=0,
point meta max=1
]
\addplot graphics [xmin=0.5,xmax=32.5,ymin=0.5,ymax=32.5] {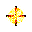};
\node[right, inner sep=0mm, text=black]
at (axis cs:-2,-2,0) {(c)};
\end{axis}
\end{tikzpicture}%
\begin{tikzpicture}[font=\small]

\begin{axis}[%
width=1.5in,
height=1.50333333333333in,
axis on top,
clip=false,
scale only axis,
xmin=0.5,
xmax=32.5,
xtick={ 1,  5, 10, 15, 20, 25, 30},
y dir=reverse,
ymin=0.5,
ymax=32.5,
ytick={ 1,  5, 10, 15, 20, 25, 30},
title={Estimated $\bb s$},
colormap={mymap}{[1pt] rgb(0pt)=(1,1,1); rgb(16pt)=(1,1,0); rgb(40pt)=(1,0,0); rgb(63pt)=(0.0416667,0,0)},
colorbar,
colorbar style={ytick={  0, 0.2, 0.4, 0.6, 0.8,   1}},
point meta min=0,
point meta max=1
]
\addplot graphics [xmin=0.5,xmax=32.5,ymin=0.5,ymax=32.5] {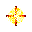};
\node[right, inner sep=0mm, text=black]
at (axis cs:-2,-2,0) {(d)};
\end{axis}
\end{tikzpicture}%
\caption{True attenuation (a) and brightness (b) components of the variable $\bf x$.
Reconstructed attenuation (c) and brightness (d) components.}
\label{lccros}
\end{figure}

True versus estimated attenuation and brightness maps for the three objects are given in Figs. 
\ref{lcsquare}, 
\ref{lccirc}, and 
\ref{lccros}. 

We see that for the convex shapes we achieve a good reconstruction while the result for the concave shape (the cross) is poorer, essentially only reconstructing the convex hull. 
This agrees with results from lightcurve inversion research \cite{calef_photometric_2006,dillon_computational_2012}.
Theoretical results in that field state that convexity of the object shape is a sufficient condition to guarantee reconstruction.

%
%



\section{Discussion}

We started by considering linear systems that result from imaging techniques which can be described with projections along rays. 
The systems could often be described as a small variation on tomographic reconstruction, at least until we included occlusion.
We demonstrated how occlusion may be handled in a computational imaging framework, by treating opacity as a case of attenuation.
It should be possible to improve on the result, therefore, by somehow adding the constraint that the attenuation can only be zero or one.
Further, one might add the constraint that when the brightness is nonzero the attenuation must be zero, and when the brightness is zero the attenuation must be one.
However we can also view the ability to handle non-opaque objects as a strength of the approach.
Also, allowing the attenuation and brightness to be independent variables allows additional types of imaging systems to be designed, as the simulations demonstrated.


As mentioned in the introduction, we restricted our analysis to the high-frequency limit.
A matrix which describes diffracting light would not in principle be impossible, resulting in a generalization of the diffraction tomography techniques \cite{devaney_reconstructive_1986} rather than ray-based tomography, but that matrix would also be much denser than one describing rays, requiring vastly more memory resources, and limiting application to smaller object sizes or lower resolution.

Generally we find that to implement much larger, not to mention three-dimensional, systems would require efficient implementation of the forward model. 
For example the Radon transform may be viewed as a particular collection for a linear imaging system, which may be represented by a matrix, but it's more efficient to simply compute it directly by summing over pixels along rays. 
Such a technique may be very useful here as well, as the memory requirements of the system as implemented tend to be the limiting factor in problem size.

We demonstrated an approach to reconstruct images of objects given lightcurve measurements.
We used a two-dimensional simulation and assumed Lambertian reflectance. 
The classic approach to lightcurve inversion requires the object to be convex.
The approach described here, being nonparametric, makes no such requirements but we can see in the results an apparent degradation of the result for the concave object.
We could also imagine techniques which utilize self-occlusion to perform super-resolution reconstruction of low-resolution images.


\bibliographystyle{siam}
\bibliography{zoterorefs}
\end{document}